\documentclass{zapiski}
\originfo{541}{}{}{2025}
\usepackage[T2A]{fontenc}
\usepackage[russian, english]{babel}

\usepackage{url}
\usepackage{booktabs}
\usepackage{graphicx}
\usepackage{cite}
\usepackage{enumerate}

\usepackage{cite}
\usepackage{amsmath,amssymb,amsfonts}
\usepackage{url}
\usepackage{graphicx}
\usepackage{textcomp}
\usepackage{color,soul}
\usepackage[table,dvipsnames]{xcolor}
\colorlet{shadecolor}{gray!20}
\usepackage{array}
\usepackage{booktabs}
\usepackage{multirow}
\usepackage{bm}
\usepackage{makecell}
\usepackage{amsmath}
\usepackage{mathtools}
\usepackage{bbm}
\usepackage{pifont}% http://ctan.org/pkg/pifont
\newcommand{\cmark}{\ding{51}}%
\newcommand{\xmark}{\ding{55}}%

\usepackage{subcaption}
\usepackage{algorithm}
\usepackage{algpseudocode}

\newcommand{\best}[1]{\mathbf{#1}}      % лучший
\newcommand{\second}[1]{\underline{#1}} % второй лучший

\newcommand{\metric}[3]{\ensuremath{#3{#1} \pm #2}}

\newcommand{\metricplain}[2]{\metric{#1}{#2}{}}
\newcommand{\metricbest}[2]{\metric{#1}{#2}{\best}}
\newcommand{\metricsecond}[2]{\metric{#1}{#2}{\second}}

\begin{document}
\sloppy
\title[]{Probabilistic distances-based hallucination detection in LLMs with RAG}

\author{Rodion Oblovatny\textsuperscript{*}}
\address[Rodion Oblovatny]{Markov Lab, Department of Mathematics and Computer Science, Saint-Petersburg University, Saint-Petersburg, Russia}
\email{oblovatnyi@gmail.com}

\author{Alexandra Kuleshova\textsuperscript{*}}
\address[Alexandra Kuleshova]{AI Center, Skoltech, Moscow, Russia}
\email{A.Bazarova@skoltech.ru}

\author{Konstantin Polev}
\address[Konstantin Polev]{SB AI Lab, Moscow, Russia}
\email{endless.dipole@gmail.com}

\author{Alexey Zaytsev}
\address[Alexey Zaytsev]{AI Center, Skoltech, Risk department, Sber, Moscow, Russia}
\email{A.Zaytsev@skoltech.ru}

\begingroup
\renewcommand\thefootnote{\fnsymbol{footnote}}
\footnotetext[1]{Equal contribution}
\endgroup
\renewcommand{\shortauthors}{R.~Oblovatny, A.~Kuleshova, K.~Polev, A.~Zaytsev}

\maketitle

\begin{abstract}
% We present a novel approach for detecting hallucinations in large language models (LLMs) by analyzing the probabilistic divergence between prompt and response hidden-state distributions. Counterintuitively, we find that hallucinated responses exhibit smaller deviations from their prompts compared to grounded responses, suggesting that hallucinations often arise from superficial rephrasing rather than substantive reasoning. Leveraging this insight, we propose a model-intrinsic detection method\footnote[1]{The code of the proposed method and the considered baselines is available at \url{https://anonymous.4open.science/r/halludetection-7FE6}} that uses distributional distances as principled hallucination scores, eliminating the need for external knowledge or auxiliary models. To enhance sensitivity, we employ deep learnable kernels that automatically adapt to capture nuanced geometric differences between distributions. Our approach outperforms existing baselines, demonstrating state-of-the-art performance on several benchmarks. The method remains competitive even without kernel training, offering a robust, scalable solution for hallucination detection.
Detecting hallucinations in large language models (LLMs) is critical for their safety in many applications. Without proper detection, these systems often provide harmful, unreliable answers.
In recent years, LLMs have been actively used in retrieval-augmented generation (RAG) settings. However, hallucinations remain even in this setting, and while numerous hallucination detection methods have been proposed, most approaches are not specifically designed for RAG systems.
To overcome this limitation, we introduce a hallucination detection method based on estimating the distances between the distributions of prompt token embeddings and language model response token embeddings \footnote[1]{The code of the proposed method and the considered baselines is available at \url{https://github.com/qwefep/distrib_distances4hallucinations}}. The method examines the geometric structure of token hidden states to reliably extract a signal of factuality in text, while remaining friendly to long sequences.
Extensive experiments demonstrate that our method achieves state-of-the-art or competitive performance. It also has transferability from solving the NLI task to the hallucination detection task, making it a fully unsupervised and efficient method with a competitive performance on the final task.
\end{abstract}

% пример - K2VAE

% 1. постановка и важность задачи
% Probabilistic Time Series Forecasting (PTSF) plays a crucial role in decision-making across various fields, including economics, energy, and transportation. 

% 2. почему текущие методы плохие
% Most existing methods excell at short-term forecasting, while overlooking the hurdles of Long-term Probabilistic Time Series Forecasting (LPTSF). As the forecast horizon extends, the inherent nonlinear dynamics have a significant adverse effect on prediction accuracy, and make generative models inefficient by increasing the cost of each iteration. 

% 3. вводим наш метод, объясняем на верхнем уровне, чем он хорош
% To overcome these limitations, we introduce K^2VAE, an efficient VAE-based generative model that leverages a KoopmanNet to transform nonlinear time series into a linear dynamical system, and devises a KalmanNet to refine predictions and model uncertainty in such linear system, which reduces error accumulation in long-term forecasting. 

% 4. кратко - про эксперименты
% Extensive experiments demonstrate that K^2VAE outperforms state-of-the-art methods in both short- and long-term PTSF, providing a more efficient and accurate solution

\section{Introduction}

In recent years, large language models (LLMs) have been widely adopted in many applications. However, they often generate hallucinations~---~incorrect or fabricated content that does not match real-world facts or the provided context~\cite{huang2023survey}. The latter case is of special interest, as it refers to incorrect generations in retrieval-augmented generation (RAG) settings, where LLMs rely on retrieved information to answer user queries. Since RAG is commonly used in chatbots~\cite{akkiraju2024facts} and other AI systems~\cite{practicalRAG}, detecting hallucinations in this setting is critical. Without proper detection, these systems may provide unreliable answers, which can be harmful in high-stakes fields such as law, finance, and healthcare.

While numerous hallucination detection methods have been proposed~\cite{sahoo2024comprehensive}, most approaches are not specifically designed for RAG systems. Traditional methods often examine hallucinations in isolation, without considering how generated content relates to its supporting context. However, recent research~\cite{bazarova2025hallucination} demonstrates that explicitly analysing the relationship between a model's generation and its provided context can significantly improve detection accuracy, as hallucinated answers consistently show weaker structural connections to their context in attention patterns compared to well-grounded responses.

We propose analysing this relationship in the space of LLM hidden states, as prior work has demonstrated that these states contain discriminative patterns of hallucinations~\cite{azaria-mitchell-2023-internal, sky2024androids, chen2024inside}. Specifically, we hypothesize that the extent to which the response's embeddings deviate from those of the prompt may be indicative of hallucination, as they capture both text semantics and the model's latent ``understanding'' of the input~\cite{azaria-mitchell-2023-internal}. To quantify this deviation, we employ probabilistic distances, such as Maximum Mean Discrepancy (MMD)~\cite{JMLR:v13:gretton12a}.

% \begin{figure}[h!]
%     \centering
%     \includegraphics[width=\columnwidth]{figures/figure_1_upd_v1.pdf} % 
%     \caption{MMD distance distributions with trained kernels for hallucinated (blue) and grounded (orange) responses, shown for Llama-2-7B and Mistral-7B. Dataset: RAGTruth QA.} \label{fig:distances}
% \end{figure}

Intuitively, we expect that embeddings of hallucinated examples should differ more from the context than grounded responses. Our experiments confirm this, demonstrating that distributions of hallucinated responses experience a shift in latent state space relative to the prompt.
The observed tendency for probabilistic distances to increase for hallucination-inducing samples allows us to use these distances as an effective measure of hallucinations: the greater the distance, the higher the probability of hallucination. This simple but powerful approach demonstrates high detection efficiency, as evidenced by the results of our experiments (Tables~\ref{tab:results1},~\ref{tab:results2}). In order for the score values to be comparable for sequences of different lengths and for the test to be valid in a non-asymptotic setting, we formulate the score using not the <<raw>> values of the statistics, but based on the $p-value$ values for the corresponding null distributions.

The task of detecting hallucinations in the RAG setting is very similar to the NLI task, in which a contradiction between the hypothesis and the premise arises when the hypothesis contains new information that is absent from or opposite to the premise. Our method exploits this similarity and demonstrates good transferability between these tasks(Table~\ref{tab:roc-auc-snli}), allowing it to be applied even when labeled examples of hallucinations are not available.

Our contributions are as follows: 
\begin{enumerate}
    \item We analyze the probabilistic distances between the distributions of hidden states of prompts and responses in LLM, revealing that hallucinatory responses show greater deviations from prompts compared to grounded responses. 
    \item Leveraging this observation, we propose a novel, model-intrinsic approach to hallucination detection that utilizes these distances between distributions to construct hallucination scores without requiring external knowledge bases or auxiliary models. Our procedure utilizes nonparametric Maximum Mean Discrepancy distances with wild bootstrap on top to get a principled hallucination score.
    \item Leveraging the similarity between the NLI task and the hallucination detection task, we construct an unsupervised hallucination detector. Its metrics are competitive, even with a significant domain shift from NLI to hallucination detection.
    \item Extensive experiments with five datasets and three LLMs up to 15B in size demonstrate that our method achieves state-of-the-art or competitive performance in hallucination detection.
\end{enumerate}

\section{Related works}

\textbf{Hallucination detection.} The problem of hallucinations in LLMs has attracted significant attention recently~\cite{zhang2023siren, huang2023survey, wang2024factuality}. Existing methods can be roughly divided into several categories.

\textit{Consistency-based} methods measure the homogeneity of multiple LLM answers to estimate uncertainty~\cite{manakul2023selfcheckgpt, chen2024inside, kuhn2023semantic, qiusemantic, nikitin2024kernel}. While these methods achieve reasonable detection performance, they require generating additional responses, making them computationally expensive. Moreover, their unsupervised nature inherently limits their effectiveness. \textit{Uncertainty-based} methods rely on LLM token probabilities to assess model's confidence~\cite{fadeeva2024fact, malinin2021uncertainty}. Although efficient and easy to implement, they struggle to fully capture complex token dependencies~\cite{yaldiz-etal-2025-design} and are affected by biased token probabilities~\cite{gallegos-etal-2024-bias}. \textit{Inner state-based} methods use hidden states or statistics from attention maps as input to lightweight classifiers~\cite{azaria-mitchell-2023-internal, sky2024androids, chuang2024lookbacklensdetectingmitigating}. 

A further limitation is that most methods overlook the RAG setting, despite it is very common in LLM applications~\cite{akkiraju2024facts}. Explicitly analyzing the relationship between the RAG context and the model's response could improve detection, as context-response relationships are known to indicate hallucinations~\cite{bazarova2025hallucination}. However, prior work focused on attention maps, leaving hidden state interactions underexplored.

% \textbf{Deep kernels.} Deep trainable kernels were originally developed to overcome key limitations in kernel-based two-sample hypothesis testing. Traditional statistical approaches, like the kernel two-sample test introduced in~\cite{JMLR:v13:gretton12a}, relied on fixed kernel functions (e.g., Gaussian or Laplacian) that often lacked the flexibility to handle complex, high-dimensional data distributions. These methods were theoretically sound but suffered from two critical drawbacks: (1) their performance heavily depended on manual kernel selection, and (2) their limited expressive power constrained their applicability to real-world problems. The introduction of deep kernel learning~\cite{liu2020learning} addressed these issues by using neural networks to learn adaptive, data-driven kernel functions, significantly enhancing the discriminative power of two-sample tests.

% This advancement has since enabled diverse applications of deep kernel methods. Researchers have successfully adapted them for tasks such as change-point detection in time series~\cite{changkernel} and artificial text detection~\cite{zhang152024detecting}. However, their potential for hallucination detection in large language models remains unexplored.

\textbf{Conclusion.} Current hallucination detection methods exhibit several limitations when applied to retrieval-augmented generation (RAG) systems. The existing approaches fail to adequately model the complex statistical relationships between prompts and responses that could significantly improve detection accuracy. While previous work has primarily analyzed attention patterns in this regard~\cite{bazarova2025hallucination}, the rich semantic information encoded in transformer hidden states was overlooked. We address this gap by introducing a novel framework that employs probabilistic distances between prompt and response hidden state distributions to construct hallucination scores. We employ the NLI task to construct a fully-unsupervised version of the method that maintains the quality of the hallucination detection task.

\section{Preliminary}
\subsection{Attention mechanism}
Modern LLMs are usually based on the self-attention mechanism \cite{vaswani2017attention}. 
This mechanism enables dynamic, context-aware token interactions by computing pairwise relevance scores across sequences. 
Each attention head operates as an independent feature detector, specializing in distinct linguistic or semantic patterns (e.g., syntactic roles or coreference)~\cite{voita2019analyzing}.

Formally, head \( h \) in layer \( \ell \) generates a representation \( \mathbf{e}_{i}^{(\ell, h)} \) for token \( i \) as follows:
\[\mathbf{e}_{i}^{(\ell, h)} = \sum_{j=1}^n \alpha_{ij}^{(\ell, h)} \mathbf{W}_V^{(\ell, h)} \mathbf{x}_j^{(\ell-1)},\]
where $\mathbf{x}_j^{(\ell-1)}$ is the hidden state of token $j$ from layer $\ell-1$ and
$$\alpha_{ij}^{(\ell, h)} = \operatorname{softmax}\left(\frac{(\mathbf{W}_Q^{(\ell, h)} \mathbf{x}_i^{(\ell-1)})^\top (\mathbf{W}_K^{(\ell, h)} \mathbf{x}_j^{(\ell-1)})}{\sqrt{d_k}}\right)$$ computes normalized attention weights,
\(\mathbf{W}_Q^{(\ell, h)}\), \(\mathbf{W}_K^{(\ell, h)}\), \(\mathbf{W}_V^{(\ell, h)}\) are learnable query, key, and value projection matrices, and \(d_k\) is the key dimension scaling factor.

% Our analysis focuses on individual attention heads rather than full layers because we find that head-level representations \(\mathbf{h}_{i}^{(\ell, h)}\) preserve more useful signals for hallucination detection. As shown in Table~\ref{tab:granularity}, using layer-level representations instead leads to worse performance~---~the factual signals appear to get diluted when we look at a coarser layer-level view.

\subsection{Probabilistic distances}
To quantify the relationship between retrieved inputs and generated responses, we propose analysing the statistical divergence between their respective hidden state distributions. This approach is motivated by the hypothesis that hallucinated responses exhibit systematically different distributional characteristics (w.r.t. to the context) from grounded responses in the model's latent space. The choice of distance metric is crucial, as it must capture meaningful semantic differences while remaining robust to high-dimensional noise and accommodate the complex geometry of transformer hidden states. Probabilistic distances between distributions offer these properties while avoiding strong parametric assumptions about the underlying data. In our work, we employ the MMD distance measure.

% \subsection{Kullback-Leibler Divergence}
% The Kullback-Leibler (KL) divergence measures distributional differences through the expectation:
% \[ D_{KL}(\alpha \| \beta) = \mathbb{E}_{x\sim\alpha}\left[\log\frac{d\alpha(x)}{d\beta(x)}\right] \]

% However, this formulation requires \(\alpha\) to be absolutely continuous with respect to \(\beta\) (\(\alpha \ll \beta\)), as it depends on the Radon-Nikodym derivative \(d\alpha/d\beta\). This becomes problematic when analyzing LLM hidden states, where empirical studies~\cite{li2018measuring, meng2022locating} demonstrate these distributions concentrate on low-dimensional manifolds. When \(\alpha\) and \(\beta\) are mutually singular (i.e., \(\alpha \perp \beta\) with disjoint supports), the KL divergence is undefined - a critical limitation for our setting where prompt and response distributions may occupy different manifolds in the high-dimensional embedding space.

\textbf{Maximum Mean Discrepancy.} The Maximum Mean Discrepancy ($\operatorname{MMD}$)~\cite{JMLR:v13:gretton12a}  is a statistical measure used to quantify the difference between two probability distributions based on samples drawn from them. It operates in a reproducing kernel Hilbert space (RKHS), where it computes the distance between the mean embeddings of the two distributions.

Given a reproducing kernel Hilbert space (RKHS) \(\mathcal{H}\) with characteristic kernel \(k:\mathcal{X}\times\mathcal{X}\to\mathbb{R}\), the squared $\operatorname{MMD}$ between distributions \(\alpha\) and \(\beta\) is defined as:
\begin{align}
    \operatorname{MMD}^2_k(\alpha, \beta) = &\mathbb{E}_{\alpha\otimes\alpha} [k(x, x')]  + \mathbb{E}_{\beta\otimes\beta}[k(y, y')] - \nonumber \\
    2&\mathbb{E}_{\alpha\otimes\beta}[k(x, y)] . \nonumber
\end{align}
For two samples $X=\{x_i\}_{i = 1}^m \sim \alpha, \; Y=\{y_j\}_{j = 1}^n \sim \beta$, we consider the unbiased estimate of \(\operatorname{MMD}_k(\alpha, \beta)\): 
\begin{align}
    \widehat{\text{MMD}}^2_u&(X, Y) = \frac{1}{m(m-1)} \sum_{i \neq j}^m k(x_i, x_j) + \nonumber \\ & + \frac{1}{n(n-1)} \sum_{i \neq j}^n k(y_i, y_j) - \frac{2}{mn} \sum_{i=1}^m \sum_{j=1}^n k(x_i, y_j), \nonumber
\end{align}
in our experiments.
%converges to the true $\operatorname{MMD}_k$ value at rate \(O(\frac{1}{\sqrt{n}} + \frac{1}{\sqrt{m}})\). 

% While $\operatorname{MMD}$ avoids the curse of dimensionality through kernel embeddings, its kernel dependence leads to reduced sensitivity to fine geometric structure in low-density regions~---~a limitation we address through learned deep kernels.

\textbf{Wild bootstrap.}
Under the null hypothesis $\alpha = \beta$, the asymptotic distribution of unbiased and biased estimates of $\operatorname{MMD}$ is degenerate, which complicates the analytical calibration of test thresholds on finite samples. Consequently, $\operatorname{MMD}$ is usually combined~\cite{mmdagg} with resampling-based procedures to obtain empirical $p$-values used as a score, suitable, e.g., for hallucination detection~\cite{rabanser2019failing}.

In this work, we use the wild bootstrap procedure to approximate the zero distribution of the MMD statistic and compute $p$-values. We use wild bootstrap instead of label shuffling because wild bootstrap allows us to explicitly capture sequential dependence in the data. This is particularly important in our case, since the token-level latent representations we will use form structured sequences with strong local correlations caused by syntax, semantics, and the transformer architecture itself. Following~\cite{shao2010dependent, chwialkowski2014wild}, for two sequences $X=\{x_i\}_{i = 1}^m \sim \alpha, \; Y=\{y_j\}_{j = 1}^n \sim \beta$, we generate a sequence of bootstrap weights $W\coloneqq\{W_t\}_{t=1}^{n + m}$ according to a first-order autoregressive process,
\begin{equation}
    W_t = \rho W_{t-1} + \sqrt{1 - \rho^2}\,\varepsilon_t, 
    \quad \varepsilon_t \sim \mathcal{N}(0,1),
\end {equation}
where $\rho = \exp(-1/\ell)$ controls the temporal dependence of the weights, and $\ell$ is the bandwidth parameter. Given the weights ${W_t}$, the analogue of the MMD statistic for the wild bootstrap is obtained by weighting the kernel terms in the statistic,
\begin{align}
    \label{formula:bootstrapped-mmd}
    \widehat{\text{MMD}}^2_{w,u}&(X, Y) = \frac{1}{m(m-1)} \sum_{i \neq j}^m \tilde W_i^{(x)}\tilde W_j^{(x)} k (x_i, x_j) + \nonumber \\ & + \frac{1}{n(n-1)} \sum_{i \neq j}^n \tilde W_i^{(y)}\tilde W_j^{(y)} k(y_i, y_j) - \nonumber \\ & \frac{2}{mn} \sum_{i=1}^m \sum_{j=1}^n \tilde W_i^{(x)}\tilde W_j^{(y)} k(x_i, y_j),
\end{align}
where $W_i^{(x)} = W_i$ for $i=\overline{1,m}$, $W_j^{(y)} = W_{j + m}$ for $j=\overline{1,n}$, $\tilde W_i^{(x)} = W_i^{(x)} - \frac{1}{m}\sum_kW_k^{(x)}$ and $\tilde W_j^{(y)} = W_j^{(y)} - \frac{1}{m}\sum_kW_k^{(y)}$ are centered versions of $W_i^{(x)}$ and $W_j^{(y)}$ respectively. 

Given two samples $X=\{x_i\}_{i = 1}^m$ and $Y=\{y_j\}_{j = 1}^n$, for $b=1,\dots,B$ we generate a sequence of bootstrap weights $W^{(b)}\coloneqq \{W_t^{(b)}\}_{t=1}^{n + m}$ and calculate $M^{(b)}\coloneqq \widehat{\text{MMD}}^2_{W^{(b)},u}(X,Y)$ using formula~\eqref{formula:bootstrapped-mmd}, obtaining an approximation of the null distribution $(W_b)_{b=1}^B$ of the test statistic. Through the bootstrap procedure, we obtain a $p\text{\,-}value$:
\begin{align}
    \label{formula:p-value}
    p\text{\,-}value \coloneqq \frac{1}{B + 1}\left(1 + \sum_{b=1}^B\mathbbm{1}\left[ M^{(b)} \ge \widehat{\text{MMD}}^2_u(X,Y)\right]\right),
\end{align}
which can be used to construct a normalized score.

%Repeating this procedure gives an empirical distribution that sequentially approximates the null distribution of the test statistic.

% Through the bootstrap procedure, we obtain a $p$-value which can be used to construct a normalized score.

\section{Methodology}

\subsection{Method}

For a given prompt-response pair \([P, R]\), we propose using the MMD distance between the distributions of the hidden states of prompt tokens \(P\) and response tokens \(R\) as an indicator of hallucination: a larger distance $d(\alpha_P, \beta_R)$ indicates a higher probability of hallucination, where $\alpha_P$ and $\beta_R$ denote the corresponding distributions of hidden states of tokens.

Since we want our test to be valid in the non-asymptotic framework and comparable for sequences of different lengths, we propose using the $p\text{\,-}value$ based on the approximation of the null hypothesis using Monte Carlo approximation instead of the <<raw>> MMD value. Our experiments~\ref{subsection:boostrap req} demonstrate the advantage of this approach over the <<raw>> value of the statistic.

To achieve finer granularity, we extract hidden states from several model \textit{heads} (selected by evaluating their discriminative power on the training dataset) rather than relying on standard layer-wise embeddings. Our experiments (Table~\ref{tab:granularity}) show that utilizing head embeddings improves detection quality, suggesting that faithfulness signals become ``blurred'' at the layer level.

Formally, our hallucination scores are formed as follows. Employing a set of pre-selected heads (see head selection procedure down below) $H_{\text{selected}}$ of size $N_{opt}$, for a sample $S = [P, R]$ for each head $h \in H_{\text{selected}}$ we obtain the corresponding sequence of embeddings $[\mathbf{E}^{h}_P, \mathbf{E}^{h}_R] = [\mathbf{e}_{h, P}^1, \dots, \mathbf{e}_{h, P}^{\mathrm{len}(P)}, \mathbf{e}_{h, R}^1, \dots,  \mathbf{e}_{h, R}^{\mathrm{len}(R)}]$. Given an estimate $\hat{d}$ of the probability distance $d$, which depends on the kernel function $k$, we use the bootstrap process to obtain the empirical null distribution of the test statistic $d$ and the corresponding $p\text{\,-}value$ using formula~\ref{formula:p-value} above. The final score for head $h$ is equal to $s_k(E^{h}_P, E^{h}_R)=1 - p\text{\,-}value$. After calculating the scores for all heads, we average them:
\begin{equation}
    \mathcal{HS}_k(S) = \frac{1}{N_{opt}}\sum\limits_{h \in H_{\text{selected}}}s_k(\mathbf{E}^{h}_P, \mathbf{E}^{h}_R). \nonumber
\end{equation}

Our pipeline is briefly outlined in Algorithm~\ref{alg:head-selection-and-prediction}. Below, we describe this procedure in detail. 

\textbf{Head selection.} To select the attention heads, we adopt a procedure similar to the one proposed in~\cite{bazarova2025hallucination}. Additionally, we compute hallucination scores using either $\operatorname{MMD}$ on the hidden states, rather than the topological divergence applied to attention maps.

The procedure assumes the availability of a annotated probe dataset $V$. For a fixed head $h$, we define the separation capability: 
\begin{equation}
\begin{split}
    \Delta_h&=\frac{1}{\#\{(P, R, y)\in V| y = 1\}}\sum_{\{(P, R, y)\in V| y = 1\}}s_k(E_P^h, E_R^h) - \\
    &\frac{1}{\#\{(P, R, y)\in V| y = 0\}}\sum_{{(P, R, y)\in V| y = 0\}}}s_k(E_P^h, E_R^h),
\end{split}
\label{formula:delta-head}
\end{equation}
where $s_k$ we defined above. 
The procedure consists of three main steps:
\begin{enumerate}
    \item For each head $h$ in the model, we measure the probabilistic distances across all training samples.
    \item We rank the heads by their separation capability $\Delta_h$.
    \item From the top-performing heads, we select the optimal subset of size $N_{\mathrm{opt}}$ based on their joint performance (hallucination score as the average from the subset) measured on the validation set.
\end{enumerate}
In line with~\cite{bazarova2025hallucination}, we constrain $N_{\mathrm{opt}}$ to a maximum of $N_{\mathrm{max}} = 6$ to maintain efficiency. An important advantage is that, as we show in experiments~\ref{results:nli-transfer}, the method demonstrates competitive performance when shifting the domain from NLI tasks to hallucination detection tasks, enabling its application in scenarios where labeled examples are unavailable. 

\begin{algorithm}
\caption{Head Selection}
\label{alg:head-selection-and-prediction}

\begin{algorithmic}[1]
\Require \\
\indent Score calculator $s_k$ \\
\indent Probe set $V = \{(P_i, R_i, y_j)\}$ \\
\indent Max heads $N_{max}$

\State
\Procedure{HeadSelection}{}
    \For{each head $h_{ij}$} \Comment{Evaluate heads}
        \State Calculate $\Delta_{h_{ij}}$ according to eq.~\ref{formula:delta-head}
    \EndFor
    \State $H \gets \text{Sort}(\{h_{ij}\}, \text{key}=\Delta_{h_{ij}}, \text{descending})$
    \State $H_{\text{subset}} \gets \emptyset$, $\text{AUROC}_{\text{max}} \gets 0$, $N_{\text{opt}} \gets 1$
    \For{$N = 1$ \textbf{to} $N_{max}$}
        \State $H_{\text{subset}} \gets H_{\text{subset}} \cup \{H_N\}$
        \For{each $(P_j, R_j, y_j) \in V$}
            \State $p_j \gets \frac{N-1}{N}p_j + \frac{1}{N}s_k(P_j^{H_N}, R_j^{H_N})$
        \EndFor
        \State $\text{AUROC} \gets \text{AUROC}(\{y_j\}, \{p_j\})$
        \If{$\text{AUROC} > \text{AUROC}_{\text{max}}$}
            \State $\text{AUROC}_{\text{max}} \gets \text{AUROC}$
            \State $N_{\text{opt}} \gets N$
        \EndIf
    \EndFor
    \State \textbf{return} $H_{\text{subset}} = \{H_1, \dots, H_{N_{\text{opt}}}\}$
\EndProcedure
\State $H_{\text{selected}} \gets \text{HeadSelection}()$
\end{algorithmic}
\end{algorithm}

\section{Experiments}
\subsection{Datasets}
The proposed approach was evaluated on four datasets: RAGTruth~\cite{niu-etal-2024-ragtruth}, CoQA~\cite{reddy2019coqa}, SQuAD~\cite{Rajpurkar2016SQuAD1Q} and XSum~\cite{xsum}.
The RAGTruth dataset contains manually annotated answers for several language models in RAG setting. The dataset includes annotations for samples for several tasks; in our paper, we experiment with two subsets of this dataset corresponding to the question answering (QA)  and summarization (Summary) tasks.

As for the CoQA and SQuAD datasets, we employed the versions from the authors of the paper~\cite{bazarova2025hallucination}. The authors of this paper used questions from the original datasets, generated answers using a language model, and annotated the resulting samples automatically using GPT-4o\cite{Hurst2024GPT4oSC}.

The statistics of the considered datasets are provided in Tables~\ref{tab:data-total-stats}-\ref{tab:data-lenghts} (see Appendix\ref{app: datasets}). Note that RAGTruth consists of samples with longer contexts and responses compared to SQuAD and CoQA, thus being more representative from the real-world performance point of view. 
\subsection{Baselines}
We compare our approach with nine established baselines spanning three methodological categories. For supervised hidden state-based methods, we evaluate SAPLMA, a linear probe over LM hidden states~\cite{azaria-mitchell-2023-internal}, and an attention-pooling probe that learns weighted combinations of layer activations~\cite{sky2024androids}. The consistency-based methods include SelfCheckGPT measuring generation agreement via BERTScore~\cite{manakul2023selfcheckgpt}, semantic entropy over clustered responses~\cite{kuhn2023semantic}, and EigenScore analyzing variability in eigenspace~\cite{chen2024inside}. The uncertainty-based methods comprise self-evaluation, perplexity, and maximum entropy~\cite{fadeeva2024fact}.
\subsection{Models}
The method was evaluated on three popular open-source models: LLaMA-2-7B-chat~\cite{llama2}, Mistral-7B-Instruct-v0.1~\cite{mistral}, and LLaMA-2-13B-chat~\cite{llama2}.

\subsection{Implementation details}

For Maximum Mean Discrepancy, we used a Laplacian kernel with $l1$-norm as the kernel: \(k(x, y)=\exp(-\frac{||x-y||_{1}}{\sigma})\), where $\sigma$ is selected using median heuristics: $\sigma = \mathrm{median}(||z_i - z_j||, i<j)$ for the pooled sequence $Z=[X, Y]$. The following parameters were set for wild bootstrap: bandwidth parameter $l$ equals 1, bootstrap sample size equals 2048.

As for the baselines, we employed the following settings: for the consistency-based methods, we used $N = 20$ additional generations, following the ablation study in~\cite{manakul2023selfcheckgpt}. Hidden state-based classifiers were trained on the embeddings from the $16$-th model layer, as the prior work has shown that the middle transformer layers contain the most factuality-related information~\cite{sky2024androids, azaria-mitchell-2023-internal}.

The reported results are averaged over 5 runs with different data splits, using test sets comprising 25\% of the data and a fixed validation set size of 100.

\section{Results}

\subsection{Hallucination detection}

The results of the experiments are presented in Tables~\ref{tab:results1}-\ref{tab:results2}. We compare the method with existing state-of-the-art methods for detecting hallucinations, both those requiring labeled samples and those that are fully unsupervised. The method demonstrates competitive results on datasets with long sequences (Table~\ref{tab:results1}), achieving the best or second-best results, with the exception of the CNN/DM+Recent News dataset for the LLaMa-2-7B model.

The method demonstrates particularly strong performance on the largest of the presented models, Llama-2-13B. We also present results for the shorter sequence datasets CoQA and XSum in Table~\ref{tab:results2}. The results degrade on these datasets, but still remain the best on Llama-2-13B, which may indicate that the method is more appropriate for larger models, exploiting the fact that large models better capture the factuality signal that MMD detects during hallucination detection.
%%%%%%%%%%%%%%%%%%%%%% ROC-AUC MAIN METHOD
\begin{table}[!htbp]
\centering
\caption{ROC AUC ($\uparrow$) of hallucination detection techniques for three LLMs.
The best results for each model are highlighted in bold, and the second best are underlined. Methods requiring multiple generation are highlighted separately and separated by a line.}
\label{tab:results1}

% локально уменьшаем горизонтальные отступы и масштабируем таблицу по ширине
\begingroup
\setlength{\tabcolsep}{4pt}
\resizebox{\columnwidth}{!}{%
\begin{tabular}{l c c c c}
\toprule
Method & \makecell{Single \\ generation} & MS MARCO & \makecell{CNN/DM +\\Recent News} & SQuAD\\
\midrule
\multicolumn{5}{c}{\textbf{Mistral-7B}}\\
\midrule

Semantic entropy & \xmark &
\metricbest{0.54}{0.03} &
\metricplain{0.51}{0.04} &
\metricsecond{0.70}{0.03} \\

P(True) & \xmark &
\metricplain{0.51}{0.04} &
\metricbest{0.54}{0.03} &
\metricplain{0.44}{0.04}\\

Semantic density & \xmark &
\metricsecond{0.52}{0.04} &
\metricsecond{0.52}{0.03} &
\metricplain{0.50}{0.05}\\

EigenScore       & \xmark &
\metricbest{0.54}{0.04} &
\metricplain{0.50}{0.06} &
\metricbest{0.71}{0.04}\\
\midrule
HaloScope        & \cmark &
\metricplain{0.57}{0.08} &
\metricplain{0.51}{0.10} &
\metricsecond{0.92}{0.07}\\

LLM-Check       & \cmark &
\metricplain{0.49}{0.03} &
\metricplain{0.49}{0.03} &
\metricplain{0.50}{0.03}\\

Perplexity       & \cmark &
\metricplain{0.45}{0.01} &
\metricsecond{0.54}{0.02} &
\metricplain{0.81}{0.05}\\

Max entropy      & \cmark &
\metricsecond{0.68}{0.04} &
\metricbest{0.60}{0.07} &
\metricplain{0.75}{0.05}\\

ReDEEP           & \cmark &
\metricplain{0.54}{0.02} &
\metricplain{0.47}{0.06} &
\metricplain{0.45}{0.05}\\

\textbf{MMD(ours)}         & \cmark &
\metricbest{0.69}{0.03} &
\metricbest{0.60}{0.05} &
\metricbest{0.93}{0.01}\\

% \textbf{MMD(snli)}         & \cmark &
% \metricsecond{0.69}{0.03} &
% \metricsecond{0.59}{0.05} &
% \metricplain{0.43}{0.03} &
% \metricplain{0.86}{0.01} &
% \metricplain{0.54}{0.03}\\

% \textbf{MMD(mnli)}         & \cmark &
% \metricsecond{0.68}{0.02} &
% \metricbest{0.60}{0.05} &
% \metricplain{0.53}{0.02} &
% \metricplain{0.41}{0.06} &
% \metricplain{0.54}{0.04}\\

\toprule

\multicolumn{5}{c}{\textbf{LLaMA-2-7B}}\\
\toprule

Semantic entropy & \xmark &
\metricplain{0.53}{0.03} &
\metricplain{0.51}{0.03} &
\metricsecond{0.73}{0.03}\\

P(True) & \xmark &
\metricbest{0.57}{0.02} &
\metricbest{0.54}{0.03} &
\metricplain{0.37}{0.03}\\

Semantic density & \xmark &
\metricplain{0.49}{0.03} &
\metricplain{0.45}{0.03} &
\metricplain{0.48}{0.03}\\

EigenScore       & \xmark &
\metricsecond{0.55}{0.03} &
\metricsecond{0.53}{0.04} &
\metricbest{0.75}{0.02}\\
\midrule
HaloScope        & \cmark &
\metricplain{0.51}{0.05} &
\metricplain{0.48}{0.05} &
\metricplain{0.67}{0.04}\\

LLM-Check        & \cmark &
\metricplain{0.44}{0.02} &
\metricplain{0.49}{0.06} &
\metricplain{0.49}{0.01}\\

Perplexity       & \cmark &
\metricplain{0.54}{0.04} &
\metricplain{0.44}{0.03} &
\metricplain{0.46}{0.03}\\

Max entropy     & \cmark &
\metricbest{0.65}{0.04} &
\metricbest{0.59}{0.06} &
\metricsecond{0.73}{0.04}\\

ReDEEP           & \cmark &
\metricplain{0.54}{0.04} &
\metricplain{0.52}{0.04} &
\metricplain{0.42}{0.08}\\

\textbf{MMD(ours)}         & \cmark &
\metricsecond{0.64}{0.03} &
\metricsecond{0.53}{0.03} &
\metricbest{0.78}{0.05}\\

% \textbf{MMD(snli)}         & \cmark &
% \metricbest{0.65}{0.04} &
% \metricplain{0.55}{0.03} &
% \metricplain{0.25}{0.04} &
% \metricplain{0.54}{0.03} &
% \metricplain{0.53}{0.03}\\

% \textbf{MMD(mnli)}         & \cmark &
% \metricsecond{0.63}{0.03} &
% \metricplain{0.55}{0.03} &
% \metricplain{0.44}{0.03} &
% \metricplain{0.40}{0.03} &
% \metricplain{0.51}{0.06}\\

\midrule

\multicolumn{5}{c}{\textbf{LLaMA-2-13B}}\\
\midrule

Semantic entropy & \xmark &
\metricbest{0.57}{0.04} &
\metricbest{0.54}{0.03} &
\metricsecond{0.65}{0.03} \\

P(True) & \xmark &
\metricplain{0.55}{0.03} &
\metricbest{0.54}{0.05} &
\metricbest{0.84}{0.03} \\

Semantic density & \xmark &
\metricplain{0.52}{0.03} &
\metricsecond{0.52}{0.02} &
\metricplain{0.52}{0.02} \\

EigenScore       & \xmark &
\metricsecond{0.56}{0.04} &
\metricplain{0.47}{0.04} &
\metricplain{0.57}{0.02} \\
\midrule
HaloScope       & \cmark &
\metricplain{0.54}{0.09} &
\metricplain{0.51}{0.04} &
\metricplain{0.55}{0.02} \\

LLM-Check         & \cmark &
\metricplain{0.49}{0.06} &
\metricbest{0.56}{0.05} &
\metricplain{0.57}{0.07} \\

Perplexity       & \cmark &
\metricplain{0.54}{0.04} &
\metricplain{0.46}{0.07} &
\metricplain{0.45}{0.02} \\

Max entropy      & \cmark &
\metricsecond{0.62}{0.03} &
\metricplain{0.53}{0.06} &
\metricsecond{0.78}{0.02} \\

ReDEEP           & \cmark &
\metricsecond{0.62}{0.06} &
\metricplain{0.48}{0.05} &
\metricplain{0.48}{0.07} \\

\textbf{MMD(ours)}         & \cmark &
\metricbest{0.65}{0.03} &
\metricsecond{0.55}{0.07} &
\metricbest{0.94}{0.02} \\

% \textbf{MMD(snli)}         & \cmark &
% \metricplain{0.59}{0.06} &
% \metricplain{0.53}{0.05} &
% \metricplain{0.21}{0.04} &
% \metricbest{0.94}{0.02} &
% \metricplain{0.57}{0.03}\\

% \textbf{MMD(mnli)}         & \cmark &
% \metricsecond{0.64}{0.03} &
% \metricplain{0.53}{0.03} &
% \metricplain{0.33}{0.04} &
% \metricplain{0.43}{0.03} &
% \metricplain{0.55}{0.05}\\

\bottomrule
\end{tabular}
}% end of resizebox
\endgroup

\end{table}

%%%%%%%%%%%%%%%%%%%%%%

% %%%%%%%%%%%%%%%%%%%%%% ROC-AUC MAIN METHOD 2
\begin{table}[!htbp]
\centering
\caption{ROC AUC ($\uparrow$) of hallucination detection techniques for three LLMs for \textbf{CoQA} and \textbf{XSum} datasets. The best results for each model are highlighted in bold, and the second best are underlined. Methods requiring multiple generation are highlighted separately and separated by a line.}
\label{tab:results2}

% локально уменьшаем горизонтальные отступы и масштабируем таблицу по ширине
\begingroup
\setlength{\tabcolsep}{4pt}
\resizebox{\columnwidth}{!}{%
\begin{tabular}{l c c c}
\toprule
Method & \makecell{Single \\ generation} & CoQA & XSum\\
\midrule
\multicolumn{4}{c}{\textbf{Mistral-7B}}\\
\midrule

Semantic entropy & \xmark &
\metricbest{0.83}{0.02} &
\metricsecond{0.70}{0.03}\\

P(True) & \xmark &
\metricplain{0.61}{0.03} &
\metricplain{0.44}{0.04}\\

Semantic density & \xmark &
\metricplain{0.56}{0.02} &
\metricplain{0.50}{0.05}\\

EigenScore       & \xmark &
\metricsecond{0.74}{0.02} &
\metricbest{0.71}{0.04}\\
\midrule
HaloScope        & \cmark &
\metricsecond{0.62}{0.08} &
\metricsecond{0.92}{0.07}\\

LLM-Check       & \cmark &
\metricplain{0.60}{0.01} &
\metricplain{0.50}{0.03}\\

Perplexity       & \cmark &
\metricplain{0.54}{0.03} &
\metricplain{0.81}{0.05}\\

Max entropy      & \cmark &
\metricbest{0.73}{0.00} &
\metricplain{0.75}{0.05}\\

ReDEEP           & \cmark &
\metricplain{0.59}{0.03} &
\metricplain{0.45}{0.05}\\

\textbf{MMD(ours)}         & \cmark &
\metricbest{0.73}{0.03} &
\metricbest{0.93}{0.01}\\

% \textbf{MMD(snli)}         & \cmark &
% \metricsecond{0.69}{0.03} &
% \metricsecond{0.59}{0.05} &
% \metricplain{0.43}{0.03} &
% \metricplain{0.86}{0.01} &
% \metricplain{0.54}{0.03}\\

% \textbf{MMD(mnli)}         & \cmark &
% \metricsecond{0.68}{0.02} &
% \metricbest{0.60}{0.05} &
% \metricplain{0.53}{0.02} &
% \metricplain{0.41}{0.06} &
% \metricplain{0.54}{0.04}\\

\toprule

\multicolumn{4}{c}{\textbf{LLaMA-2-7B}}\\
\toprule

Semantic entropy & \xmark &
\metricbest{0.76}{0.01} &
\metricsecond{0.73}{0.03}\\

P(True) & \xmark &
\metricplain{0.53}{0.03} &
\metricplain{0.37}{0.03}\\

Semantic density & \xmark &
\metricplain{0.52}{0.05} &
\metricplain{0.48}{0.03}\\

EigenScore       & \xmark &
\metricsecond{0.61}{0.03} &
\metricbest{0.75}{0.02}\\
\midrule
HaloScope        & \cmark &
\metricplain{0.61}{0.04} &
\metricplain{0.67}{0.04}\\

LLM-Check        & \cmark &
\metricplain{0.60}{0.03} &
\metricplain{0.49}{0.01}\\

Perplexity       & \cmark &
\metricbest{0.74}{0.02} &
\metricplain{0.46}{0.03}\\

Max entropy     & \cmark &
\metricplain{0.65}{0.03} &
\metricsecond{0.73}{0.04}\\

ReDEEP           & \cmark &
\metricsecond{0.72}{0.04} &
\metricplain{0.42}{0.08}\\

\textbf{MMD(ours)}         & \cmark &
\metricplain{0.68}{0.04} &
\metricbest{0.78}{0.05}\\

% \textbf{MMD(snli)}         & \cmark &
% \metricbest{0.65}{0.04} &
% \metricplain{0.55}{0.03} &
% \metricplain{0.25}{0.04} &
% \metricplain{0.54}{0.03} &
% \metricplain{0.53}{0.03}\\

% \textbf{MMD(mnli)}         & \cmark &
% \metricsecond{0.63}{0.03} &
% \metricplain{0.55}{0.03} &
% \metricplain{0.44}{0.03} &
% \metricplain{0.40}{0.03} &
% \metricplain{0.51}{0.06}\\

\midrule

\multicolumn{4}{c}{\textbf{LLaMA-2-13B}}\\
\midrule

Semantic entropy & \xmark &
\metricbest{0.76}{0.04} &
\metricsecond{0.65}{0.03}\\

P(True) & \xmark &
\metricsecond{0.67}{0.02} &
\metricbest{0.84}{0.03}\\

Semantic density & \xmark &
\metricplain{0.48}{0.03} &
\metricplain{0.52}{0.02}\\

EigenScore       & \xmark &
\metricplain{0.57}{0.03} &
\metricplain{0.57}{0.02}\\
\midrule
HaloScope       & \cmark &
\metricplain{0.57}{0.03} &
\metricplain{0.55}{0.02}\\

LLM-Check         & \cmark &
\metricplain{0.57}{0.02} &
\metricplain{0.57}{0.07}\\

Perplexity       & \cmark &
\metricplain{0.62}{0.03} &
\metricplain{0.45}{0.02}\\

Max entropy      & \cmark &
\metricplain{0.66}{0.03} &
\metricsecond{0.78}{0.02}\\

ReDEEP           & \cmark &
\metricsecond{0.73}{0.02} &
\metricplain{0.48}{0.07}\\

\textbf{MMD(ours)}         & \cmark &
\metricbest{0.84}{0.03} &
\metricbest{0.94}{0.02}\\

% \textbf{MMD(snli)}         & \cmark &
% \metricplain{0.59}{0.06} &
% \metricplain{0.53}{0.05} &
% \metricplain{0.21}{0.04} &
% \metricbest{0.94}{0.02} &
% \metricplain{0.57}{0.03}\\

% \textbf{MMD(mnli)}         & \cmark &
% \metricsecond{0.64}{0.03} &
% \metricplain{0.53}{0.03} &
% \metricplain{0.33}{0.04} &
% \metricplain{0.43}{0.03} &
% \metricplain{0.55}{0.05}\\

\bottomrule
\end{tabular}
}% end of resizebox
\endgroup

\end{table}

% %%%%%%%%%%%%%%%%%%%%%%

\subsection{Transfer from Natural Language Inference task}
\label{results:nli-transfer}
\begin{figure}[!htbp]
\centering
\includegraphics[width=0.60\columnwidth]{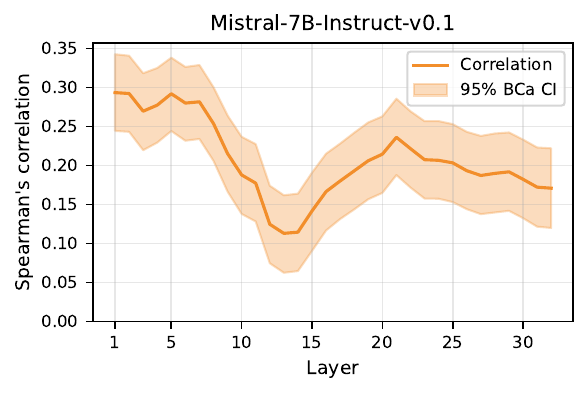}

\vspace{0.6em}

\includegraphics[width=0.60\columnwidth]{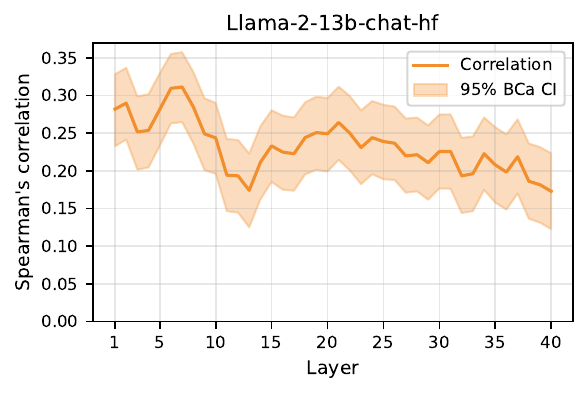}

\caption{Spearman's rank correlation coefficient between $\mathcal{HS}_k$, calculated based on LLM layer embeddings, and the NLI score of the SOTA NLI model \textbf{roberta-large-mnli}. All experiments were conducted on a held-out test set. Top: Mistral-7B-Instruct-v0.1 embeddings.
Bottom: Llama-2-13b-chat-hf embeddings.}
\label{fig:snli-spearman}
\end{figure}

The task of detecting hallucinations is very similar to the task of Natural Language Inference. We conduct experiments with head selection on the SNLI~\cite{snli} dataset and transfer to datasets for hallucination detection. We expect the method to behave similarly to natural language inference, since contradictions arise when there is a deviation from the given premise. Figure~\ref{fig:snli-spearman} shows the Spearman correlation between the resulting score and the SOTA NLI model roberta-large-mnli~\cite{liu2019roberta} on a test dataset for two models across LLM layers. The method demonstrates a strong stable correlation with the NLI score, which confirms the hypothesis that the method is sensitive to the emergence of new information in the hypothesis compared to the information presented in the premise.
Table~\ref{tab:roc-auc-snli} presents the results of transferring our method with heads selected on the NLI task to datasets for hallucination detection. The method demonstrates competitive results even with a significant domain shift from the NLI task to hallucination detection. It is important to note that in this setting, the method does not require a labeled sample with hallucinations at all.

%%%%%%%%%%%%%%%%%%%%%% ROC-AUC NLI DATASETS
\begin{table}[!htbp]
\centering
\caption{ROC AUC ($\uparrow$) of hallucination detection techniques for three LLMs during transfer MMD from SNLI dataset to the hallucination detection task.
The best results for each model are highlighted in bold, and the second best are underlined. Methods requiring multiple generation are highlighted separately and separated by a line. Comparison with methods that do not require labeled data.}
\label{tab:roc-auc-snli}

% локально уменьшаем горизонтальные отступы и масштабируем таблицу по ширине
\begingroup
\setlength{\tabcolsep}{4pt}
\resizebox{\columnwidth}{!}{%
\begin{tabular}{l c c c}
\toprule
Method & MS MARCO & \makecell{CNN/DM +\\Recent News} & SQuAD\\
\midrule
\multicolumn{4}{c}{\textbf{Mistral-7B}}\\
\midrule

Semantic entropy &
\metricbest{0.54}{0.03} &
\metricplain{0.51}{0.04} &
\metricsecond{0.70}{0.03} \\

P(True) &
\metricplain{0.51}{0.04} &
\metricbest{0.54}{0.03} &
\metricplain{0.44}{0.04} \\

Semantic density &
\metricsecond{0.52}{0.04} &
\metricsecond{0.52}{0.03} &
\metricplain{0.50}{0.05} \\

EigenScore       &
\metricbest{0.54}{0.04} &
\metricplain{0.50}{0.06} &
\metricbest{0.71}{0.04} \\
\midrule
LLM-Check       &
\metricplain{0.49}{0.03} &
\metricplain{0.49}{0.03} &
\metricplain{0.50}{0.03} \\

Perplexity       &
\metricplain{0.45}{0.01} &
\metricplain{0.54}{0.02} &
\metricsecond{0.81}{0.05} \\

Max entropy      &
\metricsecond{0.68}{0.04} &
\metricbest{0.60}{0.07} &
\metricplain{0.75}{0.05} \\

\textbf{MMD(snli)}         &
\metricbest{0.69}{0.03} &
\metricsecond{0.59}{0.05} &
\metricbest{0.86}{0.01} \\

\midrule

\multicolumn{4}{c}{\textbf{LLaMA-2-7B}}\\
\midrule

Semantic entropy &
\metricplain{0.53}{0.03} &
\metricsecond{0.51}{0.03} &
\metricsecond{0.73}{0.03} \\

P(True) &
\metricbest{0.57}{0.02} &
\metricbest{0.54}{0.03} &
\metricplain{0.37}{0.03} \\

Semantic density &
\metricplain{0.49}{0.03} &
\metricplain{0.45}{0.03} &
\metricplain{0.48}{0.03} \\

EigenScore       &
\metricsecond{0.55}{0.03} &
\metricplain{0.53}{0.04} &
\metricbest{0.75}{0.02} \\
\midrule
LLM-Check        &
\metricplain{0.44}{0.02} &
\metricplain{0.49}{0.06} &
\metricplain{0.49}{0.01} \\

Perplexity       &
\metricsecond{0.54}{0.04} &
\metricplain{0.44}{0.03} &
\metricplain{0.46}{0.03} \\

Max entropy     &
\metricbest{0.65}{0.04} &
\metricbest{0.59}{0.06} &
\metricbest{0.73}{0.04} \\

\textbf{MMD(snli)}         &
\metricbest{0.65}{0.04} &
\metricsecond{0.55}{0.03} &
\metricsecond{0.54}{0.03} \\

\midrule
\multicolumn{4}{c}{\textbf{LLaMA-2-13B}}\\
\midrule

Semantic entropy &
\metricbest{0.57}{0.04} &
\metricbest{0.54}{0.03} &
\metricsecond{0.65}{0.03} \\

P(True) &
\metricplain{0.55}{0.03} &
\metricbest{0.54}{0.05} &
\metricbest{0.84}{0.03} \\

Semantic density &
\metricplain{0.52}{0.03} &
\metricsecond{0.52}{0.02} &
\metricplain{0.52}{0.02} \\

EigenScore       &
\metricsecond{0.56}{0.04} &
\metricplain{0.47}{0.04} &
\metricplain{0.57}{0.02} \\
\midrule
LLM-Check         &
\metricplain{0.49}{0.06} &
\metricbest{0.56}{0.05} &
\metricplain{0.57}{0.07} \\

Perplexity       &
\metricplain{0.54}{0.04} &
\metricplain{0.46}{0.07} &
\metricplain{0.45}{0.02} \\

Max entropy      &
\metricbest{0.62}{0.03} &
\metricsecond{0.53}{0.06} &
\metricsecond{0.78}{0.02} \\

\textbf{MMD(snli)}         &
\metricsecond{0.59}{0.06} &
\metricsecond{0.53}{0.05} &
\metricbest{0.94}{0.02} \\
\bottomrule
\end{tabular}
}% end of resizebox
\endgroup

\end{table}

%%%%%%%%%%%%%%%%%%%%%%

\section{Ablation study}

\subsection{What about another NLI datasets?}
We additionally conduct experiments on transfer from different NLI datasets to the hallucination detection task. The results are presented in Table~\ref{tab:roc-auc-nli-datasets}. This table shows that the results can vary significantly across different datasets, suggesting that the selection of the NLI dataset is important for certain specific domains of the hallucination detection task. We assume that this depends on the structure of the NLI dataset and leave the question of its selection for a specific domain for future work.
% %%%%%%%%%%%%%%%%%%%%%% ROC-AUC NLI DATASETS
\begin{table}[h]
\centering
\caption{ROC AUC of MMD for three LLMs during transfer from head selection on the NLI dataset for different datasets. The best results for each model are highlighted in bold.}
\label{tab:roc-auc-nli-datasets}

% локально уменьшаем горизонтальные отступы и масштабируем таблицу по ширине
\begingroup
\setlength{\tabcolsep}{4pt}
\resizebox{\columnwidth}{!}{%
\begin{tabular}{l c c c c}
\toprule
NLI Dataset & MS MARCO & \makecell{CNN/DM +\\Recent News} & SQuAD & XSum\\
\midrule
\multicolumn{5}{c}{\textbf{Mistral-7B}}\\

\midrule
\textbf{SNLI}~\cite{snli}         &
\metricbest{0.69}{0.03} &
\metricplain{0.59}{0.05} &
\metricbest{0.86}{0.01} &
\metricbest{0.54}{0.03}\\

\textbf{ANLI(round 1)}~\cite{anli}         &
\metricplain{0.68}{0.04} &
\metricplain{0.58}{0.06} &
\metricplain{0.85}{0.03} &
\metricbest{0.54}{0.03}\\

% \textbf{MMD(anli round 2)}         &
% \metricsecond{0.63}{0.04} &
% \metricsecond{0.56}{0.05} &
% \metricplain{0.77}{0.03} &
% \metricplain{0.52}{0.03}\\

% \textbf{MMD(anli round 3)}         &
% \metricsecond{0.63}{0.02} &
% \metricsecond{0.59}{0.06} &
% \metricsecond{0.87}{0.02} &
% \metricplain{0.50}{0.02}\\

% \textbf{MMD(docnli)}         &
% \metricsecond{0.67}{0.02} &
% \metricsecond{0.60}{0.04} &
% \metricsecond{0.36}{0.05} &
% \metricplain{0.54}{0.05}\\

\textbf{MNLI}~\cite{mnli}         &
\metricplain{0.68}{0.02} &
\metricbest{0.60}{0.05} &
\metricplain{0.41}{0.06} &
\metricbest{0.54}{0.04}\\

\toprule

\multicolumn{5}{c}{\textbf{LLaMA-2-7B}}\\
\midrule
\textbf{SNLI}~\cite{snli}         &
\metricbest{0.65}{0.04} &
\metricbest{0.55}{0.03} &
\metricplain{0.54}{0.03} &
\metricbest{0.53}{0.03}\\

\textbf{ANLI(round 1)}~\cite{anli}         &
\metricplain{0.59}{0.03} &
\metricbest{0.55}{0.04} &
\metricbest{0.55}{0.04} &
\metricplain{0.47}{0.05}\\

% \textbf{MMD(anli round 2)}         &
% \metricsecond{0.59}{0.02} &
% \metricsecond{0.54}{0.03} &
% \metricplain{0.61}{0.04} &
% \metricplain{0.46}{0.03}\\

% \textbf{MMD(anli round 3)}         &
% \metricsecond{0.57}{0.02} &
% \metricsecond{0.54}{0.04} &
% \metricplain{0.54}{0.04} &
% \metricplain{0.44}{0.04}\\

% \textbf{MMD(docnli)}         &
% \metricbest{0.64}{0.03} &
% \metricsecond{0.56}{0.02} &
% \metricplain{0.48}{0.05} &
% \metricplain{0.52}{0.06}\\

\textbf{MNLI}~\cite{mnli}         &
\metricplain{0.63}{0.03} &
\metricbest{0.55}{0.03} &
\metricplain{0.40}{0.03} &
\metricplain{0.51}{0.06}\\

\midrule
\multicolumn{5}{c}{\textbf{LLaMA-2-13B}}\\

\midrule
\textbf{SNLI}~\cite{snli}         &
\metricplain{0.59}{0.06} &
\metricplain{0.53}{0.05} &
\metricbest{0.94}{0.02} &
\metricbest{0.57}{0.03}\\

\textbf{ANLI(round 1)}~\cite{anli}         &
\metricplain{0.63}{0.04} &
\metricbest{0.55}{0.05} &
\metricplain{0.92}{0.02} &
\metricplain{0.54}{0.05}\\

% \textbf{MMD(anli round 2)}         &
% \metricbest{0.64}{0.02} &
% \metricsecond{0.55}{0.05} &
% \metricbest{0.92}{0.02} &
% \metricplain{0.55}{0.07}\\

% \textbf{MMD(anli round 3)}         &
% \metricsecond{0.57}{0.05} &
% \metricplain{0.52}{0.05} &
% \metricbest{0.84}{0.01} &
% \metricplain{0.47}{0.04}\\

% \textbf{MMD(docnli)}         &
% \metricbest{0.63}{0.03} &
% \metricplain{0.52}{0.03} &
% \metricplain{0.40}{0.03} &
% \metricplain{0.54}{0.05}\\

\textbf{MNLI}~\cite{mnli}         &
\metricbest{0.64}{0.03} &
\metricplain{0.53}{0.03} &
\metricplain{0.43}{0.03} &
\metricplain{0.55}{0.05}\\

\bottomrule
\end{tabular}
}% end of resizebox
\endgroup

\end{table}

%%%%%%%%%%%%%%%%%%%%%%

\subsection{Is bootstrap actually necessary?}
\label{subsection:boostrap req}
We compare our method with a similar approach, but in which we do not perform bootstrapping, but use the «raw» value of the MMD statistic to detect hallucinations. The results are presented in Table~\ref{tab:bootstrap}. As can be seen from the table, in most cases the bootstrap approach outperforms the «raw» statistic value, demonstrating more stable results. One of the advantages of using the bootstrap approach is that the resulting score is normalized from zero to one, while the «raw» statistic value can be within any range.

%%%%%%%%%%%%%%%%%%%%%% BOOTSTRAP
\begin{table}[h]
\centering
\caption{ROC-AUC scores for hallucination detection comparing versions with and without bootstrapping.}

\begingroup
\setlength{\tabcolsep}{4pt}
\resizebox{\columnwidth}{!}{%
\begin{tabular}{l c c c c c}
\toprule
Version & MS MARCO & \makecell{CNN/DM +\\Recent News} & CoQA & SQuAD & XSum\\
\midrule
\multicolumn{6}{c}{\textbf{Mistral-7B}}\\
\midrule
Bootstrapped         &
\metricbest{0.69}{0.03} &
\metricbest{0.60}{0.05} &
\metricbest{0.73}{0.03} &
\metricbest{0.93}{0.01} &
\metricplain{0.59}{0.04}\\

Raw         &
\metricplain{0.67}{0.02} &
\metricplain{0.53}{0.06}  &
\metricplain{0.70}{0.03} &
\metricplain{0.88}{0.01} &
\metricbest{0.61}{0.03}\\
\midrule

\multicolumn{6}{c}{\textbf{LLaMA-2-7B}}\\
\midrule

Bootstrapped &
\metricbest{0.64}{0.03} &
\metricbest{0.53}{0.03} &
\metricplain{0.68}{0.04} &
\metricbest{0.78}{0.05} &
\metricplain{0.55}{0.06}\\

Raw &
\metricplain{0.55}{0.03} &
\metricplain{0.46}{0.03} &
\metricbest{0.74}{0.02} &
\metricplain{0.76}{0.05} &
\metricbest{0.56}{0.02} \\

\midrule

\multicolumn{6}{c}{\textbf{LLaMA-2-13B}}\\
\midrule
Bootstrapped &
\metricbest{0.65}{0.03} &
\metricbest{0.55}{0.07} &
\metricplain{0.84}{0.03} &
\metricbest{0.94}{0.02} &
\metricbest{0.60}{0.02}\\

Raw &
\metricplain{0.56}{0.04} &
\metricplain{0.48}{0.03} &
\metricbest{0.85}{0.04} &
\metricplain{0.92}{0.02} &
\metricplain{0.59}{0.07}\\

\bottomrule
\end{tabular}
}% end of resizebox
\endgroup
\label{tab:bootstrap}
\end{table}

%%%%%%%%%%%%%%%%%%%%%%

\subsection{What about other approaches to bootstrapping?}
We conducted additional experiments with alternative approaches to bootstrapping under $H_0$: wild bootstrap with different methods of weight sampling, as well as label shuffling, which involves shuffling samples in the pooled sequence $Z=[X, Y]$ via permutation without replacement. 
The results are presented in Table~\ref{tab:roc-auc-hallucination-mmd-bootstrap-types}. 
As can be seen from the table, wild bootstrap with AR weights consistently performs better than other methods, suggesting that this approach to weight sampling allows for accurate accounting of the autocorrelation of sample instances and is optimal when Maximum Mean Discrepancy is applied to samples of text token embeddings.

%%%%%%%%%%%%%%%%%%%%%% ROC-AUC DIFFERENT BOOTSTRAP PROCESSES
\begin{table}[h]
\centering
\caption{ROC AUC ($\uparrow$) of MMD-based hallucination detection techniques for three LLMs for different bootstrap types.
The best results for each model are highlighted in bold, and the second best are underlined.}
\label{tab:roc-auc-hallucination-mmd-bootstrap-types}

\begingroup
\setlength{\tabcolsep}{4pt}
\resizebox{\columnwidth}{!}{%
\begin{tabular}{l c c c c c}
\toprule
Bootstrap type & MS MARCO & \makecell{CNN/DM +\\Recent News} & SQuAD \\
\midrule
\multicolumn{4}{c}{\textbf{Mistral-7B}}\\
\midrule

\textbf{Label shuffling}         &
\metricplain{0.67}{0.02} &
\metricsecond{0.59}{0.04} &
\metricsecond{0.92}{0.02}\\

\textbf{Wild with iid normal weights}         &
\metricsecond{0.68}{0.02} &
\metricsecond{0.59}{0.04} &
\metricbest{0.93}{0.01}\\

\textbf{Wild with iid rademacher weights}         &
\metricplain{0.67}{0.03} &
\metricsecond{0.59}{0.04} &
\metricbest{0.93}{0.02}\\

\textbf{Wild with AR(1) weights (ours)}         &
\metricbest{0.69}{0.03} &
\metricbest{0.60}{0.05} &
\metricbest{0.93}{0.01}\\

\toprule

\multicolumn{4}{c}{\textbf{LLaMA-2-7B}}\\
\toprule

\textbf{Label shuffling}         &
\metricsecond{0.62}{0.02} &
\metricbest{0.56}{0.03} &
\metricplain{0.65}{0.05}\\

\textbf{Wild with iid normal weights}         &
\metricbest{0.64}{0.03} &
\metricsecond{0.54}{0.03} &
\metricplain{0.72}{0.03}\\

\textbf{Wild with iid rademacher weights}         &
\metricbest{0.64}{0.03} &
\metricsecond{0.54}{0.05} &
\metricsecond{0.75}{0.07}\\

\textbf{Wild with AR(1) weights (ours)}         &
\metricbest{0.64}{0.03} &
\metricplain{0.53}{0.03} &
\metricbest{0.78}{0.05} \\

\midrule

\multicolumn{4}{c}{\textbf{LLaMA-2-13B}}\\
\midrule

\textbf{Label shuffling}         &
\metricsecond{0.63}{0.04} &
\metricsecond{0.52}{0.04} &
\metricsecond{0.94}{0.02}\\

\textbf{Wild with iid normal weights}         &
\metricbest{0.65}{0.04} &
\metricsecond{0.52}{0.05} &
\metricbest{0.95}{0.02}\\

\textbf{Wild with iid rademacher weights}         &
\metricbest{0.65}{0.04} &
\metricplain{0.51}{0.05} &
\metricsecond{0.94}{0.02}\\

\textbf{Wild with AR(1) weights (ours)}         &
\metricbest{0.65}{0.03} &
\metricbest{0.55}{0.07} &
\metricsecond{0.94}{0.02} \\

\bottomrule
\end{tabular}
}% end of resizebox
\endgroup

\end{table}

\subsection{Do we need to consider head-level embeddings?}
To evaluate the necessity of fine-grained head selection, we compared our head-level MMD approach against a layer-level variant. For fair comparison, we adapted the selection procedure from Algorithm~\ref{alg:head-selection-and-prediction} to identify optimal layers using the same training data. Results in Table~\ref{tab:granularity} reveal that head-wise embeddings consistently outperform or performs on par with layer-wise ones, achieving higher ROC-AUC in the vast majority of cases.

The advantage is most obvious on the SQuAD dataset, where head-level MMD exceeds layer-level performance by over 0.2 ROC-AUC for Llama-2-7B and 0.1 for Mistral-7B. This demonstrates that head-specific embeddings better preserve factuality signals, while layer aggregation appears to dilute discriminative information.

%%%%%%%%%%%%%%%%%%%%%% HEADS VS LAYERS
\begin{table}[h]
\centering
\caption{ROC-AUC scores for hallucination detection comparing head-wise and layer-wise embeddings.}

\begingroup
\setlength{\tabcolsep}{4pt}
\resizebox{\columnwidth}{!}{%
\begin{tabular}{l c c c c}
\toprule
Granularity & MS MARCO & \makecell{CNN/DM +\\Recent News} & SQuAD & XSum\\
\midrule
\multicolumn{5}{c}{\textbf{Mistral-7B}}\\
\midrule
Heads         &
\metricbest{0.69}{0.03} &
\metricbest{0.60}{0.05} &
\metricbest{0.93}{0.01} &
\metricbest{0.59}{0.04}\\

Layers         &
\metricplain{0.67}{0.02} &
\metricbest{0.60}{0.03} &
\metricplain{0.83}{0.01} &
\metricplain{0.55}{0.03}\\
\midrule

\multicolumn{5}{c}{\textbf{LLaMA-2-7B}}\\
\midrule

Heads         &
\metricbest{0.64}{0.03} &
\metricbest{0.53}{0.03} &
\metricbest{0.78}{0.05} &
\metricplain{0.55}{0.06}\\

Layers         &
\metricplain{0.63}{0.04} &
\metricplain{0.52}{0.03} &
\metricplain{0.55}{0.04} &
\metricbest{0.56}{0.05}\\

\midrule

\multicolumn{5}{c}{\textbf{LLaMA-2-13B}}\\
\midrule
Heads         &
\metricbest{0.65}{0.03} &
\metricbest{0.55}{0.07} &
\metricbest{0.94}{0.02} &
\metricbest{0.60}{0.02}\\

Layers         &
\metricplain{0.58}{0.03} &
\metricplain{0.52}{0.06} &
\metricplain{0.92}{0.02} &
\metricplain{0.51}{0.03}\\

\bottomrule
\end{tabular}
}% end of resizebox
\endgroup
\label{tab:granularity}
\end{table}

%%%%%%%%%%%%%%%%%%%%%%

%%%%%%%%%%%%%%%%%%%%%%
\subsection{How large should the probe sets be? }
\begin{figure}[!htbp]
\centering

% ---------- Row (a) ----------
\begin{minipage}{\textwidth}
    \centering
    \includegraphics[width=0.45\textwidth]{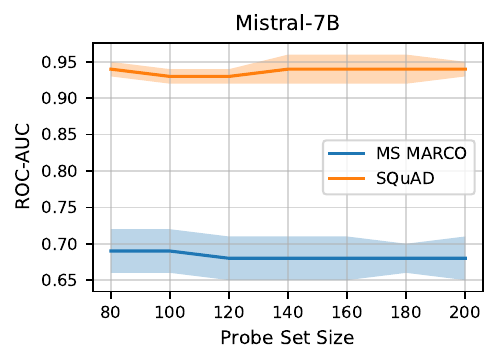}
    \hfill
    \includegraphics[width=0.45\textwidth]{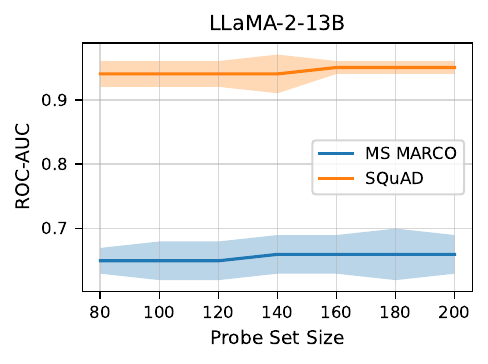}

    \phantomcaption
    \label{fig:a}
    \caption*{\textbf{a)}}
\end{minipage}

\vspace{0.5em}

% ---------- Row (b) ----------
\begin{minipage}{\textwidth}
    \centering
    \includegraphics[width=0.45\textwidth]{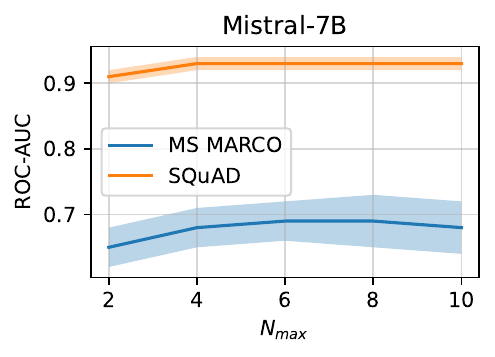}
    \hfill
    \includegraphics[width=0.45\textwidth]{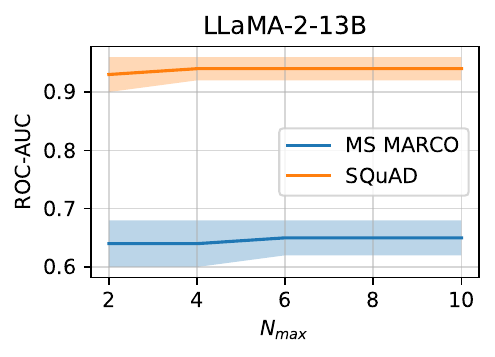}

    \phantomcaption
    \label{fig:b}
    \caption*{\textbf{b)}}
\end{minipage}

\caption{ROC-AUC performance on Mistral-7B(left) and Llama-2-13B(right) across \textbf{a)} different numbers of selected attention heads ($N_{max}$) and \textbf{b)} different probe set sizes with fixed $N_{max}=6$.}
\label{fig:roc-auc-n_sites_val_size}
\end{figure}

To evaluate the sensitivity of our method to probe set size, we conducted a study of the dependence of the results on this factor. The results presented in Table~\ref{fig:roc-auc-n_sites_val_size}a) confirm that the method is practically independent of probe set size and maintains quality even with smaller sizes.

\subsection{How many heads are needed?}
We conducted a study of the dependence of the method's performance on the number of heads $N_{max}$. The results are shown in Figure~\ref{fig:roc-auc-n_sites_val_size}b). It can be seen that the method does not depend significantly on the number of heads, and $N_{max}=6$ is the optimal number. A larger number of heads can lead to signal noise, while a smaller number can lead to insufficient robustness.

\section{Conclusion}
We propose a novel method for detecting hallucinations based on maximum mean divergence ($\operatorname{MMD}$), which measures the divergence between the distributions of the query and response hidden states.
To improve detection quality, we use attention-head embeddings rather than layer embeddings, as our ablation studies confirm that layer-level representations blur factuality signals. In addition, we use bootstrapping approaches to account for the autoregressiveness of text tokens. We further demonstrate that the method does not require a large amount of labeled data in its standard setting, nor does it require a large number of heads to capture the factuality signal. 

Notably, the method has a strong correlation with the solution to the NLI task and demonstrates good transferability with a strong shift to the hallucination detection domain, allowing the method to operate in a fully unsupervised mode.

Our method achieves state-of-the-art or competitive results on various tests, especially on larger models. Our results establish a simple but effective geometric approach to hallucination detection based on statistical divergence of latent representations.

\bibliography{references}
\bibliographystyle{amsplain}

\appendix

% \begin{figure*}[ht]
%     \centering
%     \begin{subfigure}{0.45\textwidth}
%         \centering
%         \includegraphics[width=\linewidth]{figures/sd2ot.png}
%         \caption*{a)}
%     \end{subfigure}
%     \hfill
%     \begin{subfigure}{0.46\textwidth}
%         \centering
%         \includegraphics[width=\linewidth]{figures/sd2mmd.png}
%         \caption*{b)}
%     \end{subfigure}
%     \caption{Relative change for convergence \(SD_{c, \epsilon} \rightarrow OT, \epsilon \rightarrow 0\) and \(SD_{c, \epsilon} \rightarrow MMD_{-c},\epsilon \rightarrow +\infty\) for \(c(x,y)=||x-y||_{\infty}\), Llama-2-7b-chat-hf model, RagTruthQA dataset. \textbf{(a)} \(\frac{|SD_{\epsilon, c} - OT_{c}|}{OT_{c}}\), \textbf{(b)} \(\frac{|SD_{\epsilon, c} - MMD_{-c}|}{MMD_{-c}}\)}
%     \label{fig:convsd}
% \end{figure*}
\section{Datasets}
\label{app: datasets}
Tables~\ref{tab:data-total-stats} and~\ref{tab:data-lenghts} provide comprehensive statistics for the considered datasets. Table~\ref{tab:data-total-stats} presents the distribution of hallucinated versus grounded examples, while Table~\ref{tab:data-lenghts} details the average prompt and response lengths across datasets.

The data reveals distinct characteristics among the datasets: CoQA and SQuAD contain samples with relatively short model responses, whereas RAGTruth QA and RAGTruth Summ present more challenging cases with both lengthy prompts and detailed responses. This increased complexity makes RAGTruth particularly valuable for evaluating method performance under conditions that closely resemble real-world applications.

All methods were tested on isolated subsets of the datasets, which were split into training and test subsets following the authors of the original RAGTruth paper~\cite{niu-etal-2024-ragtruth} and the annotated versions of the SQuAD and CoQA datasets from~\cite{bazarova2025hallucination}.

\begin{table}[ht]
\centering
% \footnotesize
\caption{
    Dataset statistics showing the number of hallucinated (Hal.) and non-hallucinated (Grounded) examples for each model.}\label{tab:data-total-stats}
\resizebox{\textwidth}{!}{%
\begin{tabular}{lcccccccc}
\toprule
\textbf{Model} & \multicolumn{2}{c}{\makecell{\textbf{RAGTruth} \\ \textbf{QA}}} & \multicolumn{2}{c}{\makecell{\textbf{RAGTruth} \\ \textbf{Summ}}} & \multicolumn{2}{c}{\textbf{SQuAD}} & \multicolumn{2}{c}{\textbf{CoQA}} \\
\midrule
$\#$ & Hal. & Grounded & Hal. & Grounded & Hal. & Grounded & Hal. & Grounded  \\
\midrule
Mistral-7B & 378 & 561 & 615 & 325 & 311 & 389 & 776 & 776  \\
LLaMA-2-7B & 497 & 474 & 434 & 509 & 357 & 235 & 375 & 375  \\
LLaMA-2-13B & 396 & 588 & 276 & 623 & 314 & 436 & 279 & 384  \\
\bottomrule
\end{tabular}
}
\end{table}

\begin{table}[h]
\centering
\footnotesize
\caption{
    Dataset statistics. Average prompt length (Prompt) and response length  (Response) for all models.
}\label{tab:data-lenghts}
\resizebox{\textwidth}{!}{%
\begin{tabular}{lcccccccc}
\toprule
\textbf{Model} & \multicolumn{2}{c}{\makecell{\textbf{RAGTruth} \\ \textbf{QA}}} & \multicolumn{2}{c}{\makecell{\textbf{RAGTruth} \\ \textbf{Summ}}} & \multicolumn{2}{c}{\textbf{SQuAD}} & \multicolumn{2}{c}{\textbf{CoQA}} \\
\midrule
Length & Prompt & Response & Prompt & Response & Prompt & Response & Prompt & Response  \\
\midrule
Mistral-7B & 431 & 143 & 814 & 156 & 256 & 17 & 523 & 9  \\
LLaMA-2-7B & 446 & 267 & 848 & 152 & 258 & 34 & 546 & 10  \\
LLaMA-2-13B & 447 & 213 & 784 & 142 & 292 & 17 & 548 & 10  \\
\bottomrule
\end{tabular}
}
\end{table}

\section{Computational Resources}
All experiments were run on a machine equipped with one NVIDIA A100 GPU (40 GB) and one NVIDIA L100 GPU (96 GB).

\end{document}